\documentclass[letterpaper]{article}
\usepackage{aaai}
\usepackage{ulem}
\usepackage{times}
\usepackage{helvet}
\usepackage{graphicx}
\usepackage{booktabs}
\usepackage{courier}
\usepackage{url}
\usepackage{balance}
\usepackage{natbib}
\usepackage{multirow}
\usepackage[dvipsnames,table]{xcolor}
\usepackage[colorlinks=false]{hyperref}
\hypersetup{
    colorlinks=true,
    linkcolor=MidnightBlue,   
    citecolor=RoyalBlue,      
    urlcolor=magenta         
}

\providecommand{\keywords}[1]
{
  {\small\textbf{Keywords:} #1}
}

\nocopyright
\makeatletter 
\def\@copyright@footer{Submitted to Arxiv}
\makeatother

\begin{document}
\title{SynDelay: A Synthetic Dataset for Delivery Delay Prediction}
\author{Liming Xu\textsuperscript{1}, Yunbo Long\textsuperscript{1} and Alexandra Brintrup\textsuperscript{1,2} \\
\textsuperscript{1}Supply Chain AI Lab, Department of Engineering, University of Cambridge, Cambridge CB3 0FS, UK \\
\textsuperscript{2}The Alan Turing Institute, London NW1 2DB, UK \\ 
{\tt \{lx249, yl892, ab702\}@cam.ac.uk}
}

\maketitle

\begin{abstract}
    \begin{quote}
    Artificial intelligence (AI) is transforming supply chain management, yet progress in predictive tasks---such as delivery delay prediction---remains constrained by the scarcity of high-quality, openly available datasets. 
    Existing datasets are often proprietary, small, or inconsistently maintained, hindering reproducibility and benchmarking.
    We present \textbf{SynDelay}, a synthetic dataset designed for delivery delay prediction. 
    Generated using an advanced generative model trained on real-world data, SynDelay preserves realistic delivery patterns while ensuring privacy.
    Although not entirely free of noise or inconsistencies, it provides a challenging and practical testbed for advancing predictive modelling.
    To support adoption, we provide baseline results and evaluation metrics as initial benchmarks, serving as reference points rather than state-of-the-art claims.
    SynDelay is publicly available through the \textbf{Supply Chain Data Hub}\footnote{https://supplychaindatahub.org/}, an open initiative promoting dataset sharing and benchmarking in supply chain AI. 
    We encourage the community to contribute datasets, models, and evaluation practices to advance research in this area.
    All code is openly accessible at \url{https://supplychaindatahub.org/}.
    \end{quote}
\end{abstract}

\keywords{Supply chain, Synthetic data, Benchmarking, Delivery delay prediction, Machine learning}

\section{Introduction}
Despite the rapid integration of artificial intelligence (AI) into supply chain management (SCM), progress in developing and benchmarking AI solutions is severely constrained by the \textit{lack} of high-quality, openly available datasets. 
In contrast, fields such as computer vision and natural language processing  have advanced dramatically over the past fifteen years, largely driven by large-scale, standardised datasets such as MNIST \citep{lecun1998gradient}, ImageNet \citep{deng2009imagenet}, COCO \citep{lin2014microsoft}, SQuAD \citep{rajpurkar2016squad}, and Wikipedia-based corpora.
These well-curated datasets facilitate reproducible research, enable fair model comparisons, and provide consistent benchmarks that accelerate methodological advances.

In SCM, however, such infrastructure remains largely \textit{absent}.
Existing datasets are often proprietary, small-scale, or poorly maintained, limiting reproducibility and hindering reliable benchmarking for supply chain tasks such as delivery delay prediction \citep{zheng2023federated}, 
demand forecasting \citep{carbonneau2008application}, and 
inventory optimisation \citep{kourentzes2020optimising}.
Addressing this gap is critical for advancing transparent, reproducible, and practically impactful AI research in SCM.

Recent release of logistics-focused datasets have begun to alleviate data scarcity.
The Amazon Delivery Dataset \citep{merchan20242021} and LaDe \citep{wu2024lade} provide real-world data for last-mile delivery research, particularly courier routing and delivery time prediction. 
Both offer rich spatio-temporal data, though only LaDe provides standardised formatting and comprehensive metadata.  
SupplyGraph \citep{wasi2024supplygraph} supports graph-based modelling of a small FMCG supply network, while DataCo\footnote{https://data.mendeley.com/datasets/8gx2fvg2k6/5} covers large-scale retail orders but suffers from noise, inconsistencies, and limited documentation.
While these datasets constitute important steps forward, they remain either limited in scope, scale, or curation quality, restricting their suitability as robust benchmarks for delivery delay prediction.

Although delivery delay prediction has been widely studied---ranging from traditional data analytics \citep{brintrup2020supply} to deep learning \citep{bassiouni2023advanced,gabellini2024deep} and federated learning \citep{zheng2023federated}---no common benchmark currently exists.
To address this gap, we introduce a \textit{synthetic} dataset---\textbf{SynDelay}---designed as a clean, large-scale, openly accessible dataset for delivery delay prediction.
SynDelay is carefully curated, preserves realistic delivery patterns, and includes baseline models with a set of evaluation metrics, providing a reproducible resource for transparent model comparison and supporting collective advancement of supply chain AI research.


\section{Dataset Generation and Description}
\label{sec:syndelay}
\begin{figure*}[h!]
    \centering
    \includegraphics[width=\linewidth]{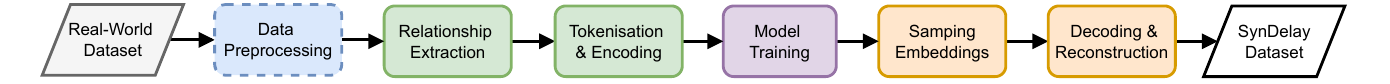}
    \caption{Workflow for generating the SynDelay dataset.}
    \label{fig:data_generation_workflow}
\end{figure*}
In this study, we leverage synthetic data to address the persistent challenge of data scarcity in SCM, particularly for predictive tasks such as delivery delay forecasting.
Synthetic data are \textit{artificially} generated to \textit{replicate} the statistical properties of real-world datasets \citep{jordon2022synthetic}.
They offer a valuable solution when real data are scarce, sensitive, or costly to obtain, providing privacy protection and enabling secure data sharing. 
Their applications have extended multiple domains beyond computer science, including medicine and healthcare \citep{chen2021synthetic,giuffre2023harnessing,koetzier2024generating}, finance \citep{assefa2020generating}, and supply chains \citep{long2025llmtablogic}.

\subsection{Data Generation}
\label{sec:data_generation}
Modern generative models, including diffusion models and large language models (LLMs), have achieved human-like performance across modalities such as images, text, audio, and video. 
However, supply chain data are predominantly \textit{tabular}, characterised by heterogeneous column types, complex inter-column dependencies, missing values, class imbalances, and temporal or integrity constraints. 
These properties present significant challenges for existing approaches when directly applied to generate high-quality synthetic tabular data \citep{long2025llmtablogic}.

To address these challenges, we develop a lightweight framework specifically designed for generating supply chain tabular data, balancing computational efficiency with data fidelity.
The framework integrates LLM-based reasoning into a score-based diffusion model operating in the latent space, enabling the generation of synthetic datasets that preserves both statistical properties and inter-column logical relationships.

Using this approach, we generated the SynDelay dataset. 
The data generation process comprises seven steps, as illustrated in \autoref{fig:data_generation_workflow}.
First, the raw data undergo preprocessing, including cleansing, noise removal, deduplication, and column selection. 
Next, an LLM infers and extracts inter-column relationships, which---together with the processed data---are tokenised, encoded, and fed into the diffusion model for training. 
The trained model is then used to sample latent \textit{embeddings}, which are decoded and reconstructed into the final synthetic dataset. 
For more technical details, see \citet{long2025llmtablogic}.

\subsection{Statistics, Variables and Metadata}
\begin{table}[t!]
\centering
\caption{Statistics of the original and synthetic datasets.}
\label{tab:dataset_statistics}
\footnotesize
\setlength{\tabcolsep}{2pt} 
    \begin{tabular}{c | c c c c c}
    \hline\rule{0pt}{1em}
    \textbf{Dataset} & \textbf{\# Rows} & \textbf{\# Vars (Num/Cat)} & \textbf{\# Classes (0/1/2)} \\
    \hline\hline\rule{0pt}{1em} 
    Original & 180,519 & 53 (24/16) & 33,753/43,366/103,400 \\
    SynDelay & 155,488 & 41 (22/12) & 29,055/36,724/89,709 \\
    \hline
\end{tabular}
\end{table}

The SynDelay dataset was generated by preprocessing a large-scale real-world delivery dataset (hereafter referred to as the \textit{Original} dataset) and feeding it into the pipeline.
Summary statistics for both the original and synthetic datasets are provided in \autoref{tab:dataset_statistics}.
The original dataset comprises 180,519 rows and 54 columns, including numerical, categorical, datetime, text variables, along with a delivery outcome column taking values (i.e., classes or labels) \{0, 1, 2\}, corresponding to \textit{early}, \textit{on-time}, and \textit{delayed} deliveries, respectively.
After preprocessing---which included noise reduction, handling missing values, deduplication, and selecting features suitable for diffusion-based modelling---the dataset was reduced to 41 variables plus the delivery outcome column.

The trained model can theoretically generate an unlimited number of samples from the learned latent space and reconstruct them as tabular data. 
For this study, we produced approximately 150k synthetic rows to balance dataset size, diversity, and computational efficiency. 
A companion metadata file, detailing the type and description of each variable, is also provided. 
The dataset is publicly available at \url{https://supplychaindatahub.org/}.

\section{Baselines and Evaluation Metrics}
\label{sec:baselines_and_metrics}
Building on the dataset introduced earlier, this section presents the remaining two key elements for benchmarking: baselines models and evaluation metrics.
As shown in \autoref{tab:dataset_statistics}, SynDelay's target variable comprises three imbalanced classes, framing this prediction problem as a multi-class imbalanced classification task.

To capture a broad performance spectrum, we evaluate five baselines. 
First, trivial models establish lower bounds: random guess, which assigns classes \textit{uniformly} at random, and ZeroR (ZeroRule), which always predicts the majority class (delayed delivery). 
Second, we consider ensemble classifiers widely adopted in tabular learning: Random Forest \citep{breiman2001random}, a bagging-based method that reduces variance through bootstrap aggregation; XGBoost \citep{chen2016xgboost} and CatBoost \citep{prokhorenkova2018catboost}, boosting-based methods that iteratively reduce bias by focusing on hard-to-predict samples. 
These ensemble methods are well-suited for mixed feature types, nonlinear relationships, and class imbalance.

Performance is measured using both aggregate metrics (Accuracy, Macro F1, Weighted F1 (Micro F1)) and per-class metrics (Precision, Recall, and F1) with emphasis on delayed deliveries---the majority class (Class 2), which represents roughly 60\% of the data. Together, these baselines and metrics establish a robust framework for benchmarking delivery delay prediction.

\section{Baseline Results}
\label{sec:results}
\begin{table*}
\centering
\caption{Prediction results of the five baseline models, reported as mean $\pm$ standard deviation over \textit{ten} independent runs. 
The top three results are highlighted in \textit{gray}, with darker shading indicating better performance.
For all metrics, \textit{higher} is better.}
\label{tab:baseline_results}
\small{
    \begin{tabular}{cc|ccccc} 
    \hline
    \rule{0pt}{1em} 
    & \textbf{Metrics}  & \textbf{Random}  & \textbf{ZeroRule}   & \textbf{Random Forest} & \textbf{XGBoost} & \textbf{CatBoost}  \\
    \hline\hline
    \rule{0pt}{1em} 
    \multirow{3}{*}{aggregate} 
      & Accuracy     & 0.3374$\pm$0.0000 & \cellcolor{gray!60}0.5770$\pm$0.0000 & 0.5724$\pm$0.0033 & \cellcolor{gray!45}0.5754$\pm$0.0019 & \cellcolor{gray!15}0.5749$\pm$0.0016  \\
      & Macro F1     & 0.3184$\pm$0.0000 & 0.2439$\pm$0.0000 & \cellcolor{gray!60}0.4833$\pm$0.0034 & \cellcolor{gray!45}0.4126$\pm$0.0028 & \cellcolor{gray!15}0.3096$\pm$0.0216  \\
      & Weighted F1  & 0.3572$\pm$0.0000 & 0.4222$\pm$0.0000 & \cellcolor{gray!60}0.5601$\pm$0.0035 & \cellcolor{gray!45}0.5295$\pm$0.0015 & \cellcolor{gray!15}0.4601$\pm$0.0118  \\
    \hline\hline
    \rule{0pt}{1em} 
    \multirow{3}{*}{\shortstack{per-class \\ (Class 2)}}
      & F1           & 0.4219$\pm$0.0000 & \cellcolor{gray!60}0.7317$\pm$0.0000 & 0.6697$\pm$0.0047 & \cellcolor{gray!15}0.7123$\pm$0.0024 & \cellcolor{gray!45}0.7238$\pm$0.0037  \\
      & Precision    & 0.5819$\pm$0.0000 & 0.5770$\pm$0.0000 & \cellcolor{gray!60}0.8830$\pm$0.0030 & \cellcolor{gray!45}0.6413$\pm$0.0022 & \cellcolor{gray!15}0.5837$\pm$0.0027  \\
      & Recall       & 0.3310$\pm$0.0000 & \cellcolor{gray!60}1.0000$\pm$0.0000 & 0.5394$\pm$0.0055 & \cellcolor{gray!15}0.8011$\pm$0.0084 & \cellcolor{gray!45}0.9528$\pm$0.0174  \\
    \hline
    \end{tabular}
}
\end{table*}

The baseline models were implemented in Python using the following libraries: sklearn\footnote{https://scikit-learn.org/stable/}, xgboost\footnote{https://xgboost.readthedocs.io/en/stable/}, and catboost\footnote{https://catboost.ai/}. 
The dataset was randomly partitioned into training, validation, and test sets with a ratio of {\tt 0.8:0.1:0.1}.
To ensure a fair and unbiased benchmark, all available variables \textit{compatible} with each model were used directly after minimal transformations or encoding, without  domain-specific feature engineering.
Ensemble classifiers were run with commonly-used hyperparameters, and each experiment was repeated \textit{ten} times, reporting mean $\pm$ standard deviation to account for variability.

The evaluation results of the five baselines are presented in \autoref{tab:baseline_results}, including aggregate metrics across all classes and per-class metrics for the majority class (Class 2, delayed deliveries).
As expected, ensemble classifiers substantially outperform trivial baselines, while ZeroRule and random guess provide useful lower bounds. 
Learning-based models achieve a better balance between overall accuracy and minority-class recognition, demonstrating their effectiveness in multi-class imbalanced settings.

Considering aggregate metrics across all classes, ZeroR achieves the highest accuracy (0.5770) by always predicting the majority class (Class 2), but its very low Macro F1 (0.2439) highlights poor minority-class recognition. 
Among ensemble methods, Random Forest delivers the most balanced performance, with the highest Macro F1 (0.4833) and Weighted F1 (0.5601). 
XGBoost and CatBoost also surpass trivial baselines, but their slightly lower Macro and Weighted F1 scores underscore the difficulty of modelling this dataset.

For the majority class, ZeroR unsurprisingly yields perfect recall (1.0000) but low precision (0.5770). 
Random Forest attains the highest precision (0.8830), CatBoost achieves the highest recall (0.9528), and XGBoost provides a balanced F1 (0.7123). 
These outcomes illustrate precision–recall trade-offs in imbalanced settings and the necessity of evaluating both aggregate and per-class metrics.

Overall, results confirm that SynDelay poses a challenging and realistic benchmark: ensemble models improve substantially over trivial baselines but still yield \textit{modest} performance, reflecting the dataset’s complexity and noise. 
The experiments are intended to establish baseline references rather than state-of-the-art performance; targeted feature engineering and hyperparameter tuning would likely lead to further improvements.

\section{Discussion and Implications}
\label{sec:discussion}
This work introduces one of the first well-curated and openly available datasets in the supply chain domain that is explicitly designed for delivery delay prediction. 
Unlike prior isolated or proprietary datasets, SynDelay provides a structured, documented resource complemented by baseline models and evaluation metrics, thereby supporting more systematic and reproducible research. 
Its design reflects the imbalance, variability, and noise that are characteristic of real-world supply chain operations, providing a challenging yet realistic benchmark for predictive modelling. 
By releasing SynDelay, we also advocate for the broader adoption of \textit{benchmarking} practices in supply chain machine learning tasks, where open and secure dataset sharing is essential for cumulative progress.

At the same time, the dataset has inherent limitations. 
It is currently restricted to the retail sector and is bounded by the capabilities and biases of the underlying synthetic data generation model presented in \citep{long2025llmtablogic}. 
While it captures meaningful statistical patterns and inter-column relationships, SynDelay also contains noise and inconsistencies, which may limit certain downstream uses.
These limitations highlight the need for continuous refinement of data generation techniques and for extending future datasets to a wider range of industries and contexts.

Taken together, this work represents an \textit{early} but important step toward bridging the gap between dataset scarcity and benchmarking needs in supply chain AI.
Lasting progress will require \textit{collective} efforts: 
expanding datasets across industries, 
improving synthetic data generation techniques, and 
developing open-source models. 
SynDelay should be seen not as an endpoint but as a \textit{catalyst} for collaborative innovation in this field.
To this end, its release through the Supply Chain Data Hub marks the beginning of a broader agenda to foster dataset sharing, benchmarking, and community-driven progress in supply chain AI research.

\section{Concluding Remarks}
\label{sec:conclusion}
This paper presented SynDelay, a synthetic dataset for delivery delay prediction that captures realistic delivery patterns while preserving privacy. 
Alongside the dataset, we provided baseline results to establish initial benchmarks, highlighting both the challenges posed by SynDelay and its value as a reproducible testbed for advancing predictive modelling in supply chains. 
These benchmarks are not intended as state-of-the-art solutions but rather as reference points to guide future research and facilitate fair comparisons.

Importantly, SynDelay is hosted on the \textbf{Supply Chain Data Hub}\footnote{https://supplychaindatahub.org/}, an ongoing initiative to promote open dataset sharing and benchmarking in the supply chain AI community. 
The authors encourage researchers and practitioners to take part in this collective effort by contributing datasets, models, and evaluation practices. 
Only through such community engagement can the field establish the open resources needed to advance reproducibility, benchmarking, and the broader adoption of AI in supply chain management.

\balance

\bibliography{references}
\bibliographystyle{plainnat}
\end{document}